\documentclass[sigconf]{acmart}
\AtBeginDocument{%
  \providecommand\BibTeX{{%
    \normalfont B\kern-0.5em{\scshape i\kern-0.25em b}\kern-0.8em\TeX}}}

\copyrightyear{2022}
\acmYear{2022}
\setcopyright{rightsretained}
\acmConference[CHI '22 Extended Abstracts]{CHI Conference on Human Factors in Computing Systems Extended Abstracts}{April 29--May 5, 2022}{New Orleans, LA, USA}
\acmBooktitle{CHI Conference on Human Factors in Computing Systems Extended Abstracts (CHI '22 Extended Abstracts), April 29--May 5, 2022, New Orleans, LA, USA}
\acmDOI{10.1145/3491101.3503811}
\acmISBN{978-1-4503-9156-6/22/04}

\usepackage{tikz}
\usetikzlibrary{positioning}
\usepackage{fontawesome}
\definecolor{light-gray}{gray}{0.95}

\usepackage[hyphenbreaks]{breakurl}

\begin{document}\sloppy

\title[Fairness and Transparency in Algorithmic Decision-Making]{A Human-Centric Perspective on Fairness and Transparency in Algorithmic Decision-Making}

\author{Jakob Schoeffer}
\email{jakob.schoeffer@kit.edu}
\orcid{0000-0003-3705-7126}
\affiliation{%
  \institution{Karlsruhe Institute of Technology (KIT)}
  \country{Germany}
}


\begin{abstract}
    Automated decision systems (ADS) are increasingly used for consequential decision-making. These systems often rely on sophisticated yet opaque machine learning models, which do not allow for understanding how a given decision was arrived at. This is not only problematic from a legal perspective, but non-transparent systems are also prone to yield unfair outcomes because their sanity is challenging to assess and calibrate in the first place---which is particularly worrisome for human decision-subjects. Based on this observation and building upon existing work, I aim to make the following three main contributions through my doctoral thesis: (a) understand how (potential) decision-subjects perceive algorithmic decisions (with varying degrees of transparency of the underlying ADS), as compared to similar decisions made by humans; (b) evaluate different tools for transparent decision-making with respect to their effectiveness in enabling people to appropriately assess the quality and fairness of ADS; and (c) develop human-understandable technical artifacts for fair automated decision-making. Over the course of the first half of my PhD program, I have already addressed substantial pieces of (a) and (c), whereas (b) will be the major focus of the second half.
\end{abstract}



\maketitle

\section{Context and Motivation}
Automated decision-making has become ubiquitous in many domains such as hiring \cite{kuncel2014hiring}, bank lending \cite{townson2020ai}, grading \cite{satariano2020british}, and policing \cite{heaven2020predictive}, among others.
As automated decision systems (ADS) are used to inform increasingly high-stakes decisions, understanding their inner workings is of utmost importance---and undesirable behavior can become a problem of societal relevance.
The underlying motives of adopting ADS are manifold: they range from cost-cutting to improving performance and enabling more robust and objective decisions \cite{harris2005automated,kuncel2014hiring,Newell2015}.
One widespread assumption is that ADS can also avoid human biases in the decision-making process \cite{kuncel2014hiring}.
In fact, if properly designed, ADS can be a valuable tool for breaking out of vicious patterns of stereotyping and contributing to social equity, for instance, in the realms of recruitment \cite{chalfin2016productivity,koivunen2019understanding}, health care \cite{grote2020ethics,triberti2020third}, or financial inclusion \cite{lepri2017tyranny}.
However, ADS are typically based on artificial intelligence (AI)---particularly machine learning (ML)---techniques, which, in turn, generally rely on historical data.
If, for instance, this underlying data is biased (e.g., because certain socio-demographic groups were favored in a disproportionate way), an ADS will pick up and perpetuate existing patterns of unfairness \cite{feuerriegel2020fair}.
Prominent examples of such behavior from the recent past are race and gender stereotyping in job ad delivery on Facebook and LinkedIn \cite{Imana21a}, as well as the discrimination of black people in the realm of facial recognition \cite{buolamwini2018gender} and recidivism prediction \cite{angwin2016machine}.
To that end, in recent years, a significant body of research has been devoted to detecting and mitigating unfairness in ADS---particularly from the techno-centric perspective of the AI/ML communities \cite{barocas2018fairness}.
Yet, most of this work has focused on formalizing the concept of fairness and enforcing certain statistical equity constraints, often without explicitly taking into account the opinions, feelings, and perceptions of the people affected by such automated decisions.

Partly as a response to multiple known instances of adverse behavior by ADS in high-stakes decision-making settings, both researchers and lawmakers have started to demand that decision-subjects be presented with more information about the inner workings of algorithms \cite{srivastava2019mathematical}. 
In fact, the EU General Data Protection Regulation (GDPR)\footnote{\url{ https://eur-lex.europa.eu/eli/reg/2016/679/oj} (last accessed: October 10, 2021)}, for instance, requires the disclosure of ``the existence of automated decision-making, including [\dots] meaningful information about the logic involved [\dots]'' to data subjects.
Beyond that, however, such regulations generally remain vague and little actionable, which often results in deficient adoption, as noticed in the context of bank lending.\footnote{\url{https://algorithmwatch.org/en/poland-credit-loan-transparency/} (last accessed: October 10, 2021)}
Moreover, while the general need for transparency appears obvious, it is also understood that there exists no ``one-size-fits-all'' explanation technique that addresses all desiderata of different stakeholders of algorithmic decision-making \cite{binns2018s,dodge2019explaining,langer2021we}.
This inevitably results in certain stakeholders'---often the system designers'---goals being (implicitly or explicitly) prioritized over more vulnerable groups, such as the decision-subjects, when choosing a means to facilitate ``transparency'' of ADS \cite{mathur2019dark}.
Additionally, it remains an open question (among many others) how to evaluate the effectiveness of explanations with respect to desiderata that are somewhat difficult to measure; such as \emph{morality, ethics, responsibility} \cite{langer2021we}, and others. 

In this doctoral thesis, I aim to address selected open questions around fairness and transparency in algorithmic decision-making from the perspective of the most vulnerable stakeholders: decision-subjects.
In Section \ref{sec:related_work}, I summarize key related work and highlight specific research gaps.
Next, in Section \ref{sec:objectives}, I derive open research questions and outline how I plan to fill the previously identified gaps.
After that, in Section \ref{sec:results}, I summarize my previous results and contributions to date; and I lay out specific next steps in my dissertation in Section \ref{sec:next_steps}. To conclude, I briefly address my long-term goals.

\begin{table*}[ht]
  \caption{Overview of my dissertation-related academic work to date.}
  \label{tab:my_work}
  \centering
  \resizebox{0.8\textwidth}{!}{
  \begin{tabular}{lll}
    \toprule
    \textbf{Contribution to} & \textbf{References} & \textbf{Venue} \\
    \midrule
    \textbf{RQ1} & \citet{schoffer2021study} & \emph{TExSS Workshop @ ACM IUI 2021} \\
    & \citet{schoeffer2021perceptions} & \emph{Hawaii International Conference on System Sciences 2022 (HICSS-55)} \\
    & \citet{schoeffer2021there} & Under review at \emph{ACM CHI 2022} \\
    \midrule
    \textbf{RQ2} & \citet{schoeffer2021appropriate} & \emph{Poster @ ACM CSCW 2021} \\
    \midrule
    \textbf{RQ3} & \citet{schoffer2021ranking} & \emph{ACM COMPASS 2021} \\
    \bottomrule
  \end{tabular} 
  }
\end{table*}

\section{Key Related Work and Research Gaps}\label{sec:related_work}
In this section, I summarize key related work along the dimensions of \emph{algorithmic fairness}, \emph{explainable AI}, and \emph{perceptions of fairness and trustworthiness}, and I highlight certain research gaps that I address in my doctoral thesis.

\paragraph{Algorithmic fairness}
Fairness is a contested concept \cite{mulligan2019thing}.
\citet{mehrabi2021survey}, similarly to existing laws and regulations, define \emph{fairness} in the context of decision-making as the ``absence of any prejudice or favoritism towards an individual or a group based on their intrinsic or acquired traits.''
Generally, the (mostly AI/ML-driven) algorithmic fairness literature distinguishes individual from group fairness definitions \cite{mehrabi2021survey}.
A widely-used notion of individual fairness is \emph{fairness through awareness} (FTA) \cite{dwork2012fairness}.
FTA says that an algorithm is fair if it gives similar predictions to similar individuals.
As stated by \citet{dwork2012fairness}, the main challenge with this notion is defining an appropriate distance metric---for instance, to measure differences in qualification between individuals.
Regarding group fairness metrics, popular notions include demographic parity \cite{calders2009building} (``predictions should be independent of sensitive information'') and equalized odds/equal opportunity \cite{hardt2016equality} (``predictive error rates should be equal for all demographic groups'').
Most often, these statistical notions are then enforced as constraints during the training phase of the associated ML model (in-processing techniques \cite{mehrabi2021survey}); meaning that predictive performance (e.g., accuracy) still remains the primary objective.
There are, however, several limitations to this: first, most existing approaches to fairness-aware ADS are specifically designed for (often binary) classification tasks, whereas fairness-aware regression is still relatively under-researched. 
Furthermore, error rate-based fairness notions---as well as the whole idea of optimizing for predictive performance in ADS---rely on the availability of ground-truth labels; for instance, whether or not an applicant \emph{actually} paid back a loan.
This is a strong assumption that is often not met in real-world decision-making scenarios \cite{lakkaraju2017selective,schoffer2021ranking}.
Specifically, if ground-truth labels are not (or selectively) available \cite{lakkaraju2017selective}, then maximizing for accuracy subject to fairness constraints seems counterintuitive and is, in fact, sub-optimal \cite{kilbertus2020fair}.
Lastly, and perhaps most importantly, it has been shown that many notions of fairness are mutually incompatible \cite{chouldechova2017fair,kleinberg2016inherent}.
For instance, demographic parity will generally be at odds with individual fairness.
This implies that there \emph{cannot} be a technical fairness definition that universally works for everyone and every use case.
Hence, it is important to consider alternative fairness criteria, such as human perceptions of fairness, as well.

\paragraph{Explainable AI}

Despite being a popular topic of current research, \emph{explainable AI} (XAI) is a natural consequence of designing ADS and, as such, has been around at least since the 1980s \cite{lewis1982role}.
Its importance, however, keeps rising as increasingly sophisticated (and opaque) AI techniques are used to inform ever more consequential decisions.
Common approaches to XAI are (a) employing ``white-box'' ML models that are interpretable by design (e.g., logistic regression), or (b) providing post-hoc explanations (e.g., LIME \cite{ribeiro2016should} or SHAP \cite{lundberg2017unified}).
Explainability is not only required by law (e.g., GDPR, ECOA\footnote{Equal Credit Opportunity Act: \url{https://www.consumer.ftc.gov/articles/0347-your-equal-credit-opportunity-rights} (last accessed: October 10, 2021)}); \citet{eslami2019user}, for instance, have shown that users’ attitudes towards algorithms change when transparency is altered.
In fact, both quantity and quality of explanations matter: \citet{kulesza2013too} explored the effects of soundness and completeness of explanations on end users’ mental models and suggest, among others, that oversimplification is problematic.
Recent findings from \citet{langer2021spare}, on the other hand, suggest that in certain cases it might make sense to withhold pieces of information in order to not evoke negative reactions.
Either way, even in the presence of explanations, people sometimes rely too heavily on system suggestions \cite{bussone2015role}, a phenomenon sometimes referred to as \emph{automation bias} \cite{de2020case,Goddard2014}.
Eventually, latest research cautions that explanations can also have negative consequences---either through intentional deception of certain stakeholders \cite{chromik2019dark} or even unsuspecting downstream effects \cite{ehsan2021explainability}.
In fact, a major share of prior work in this area has been looking at issues of opaque ADS through the lens of powerful stakeholders, such as decision-makers, and how to benefit them through explanations.
There are still significant contributions to be made with respect to designing (possibly combinations of) explanations that meet the needs of more vulnerable groups (e.g., decision-subjects).
Generally, as pointed out by \citet{langer2021we}, many prior studies have gathered mixed or inconclusive empirical evidence regarding the general effectiveness of explanations---which again demands follow-up work, for instance within specific communities or subgroups of a population.
Finally, as recent XAI literature seems to have been favoring post-hoc techniques over ``white-box'' ML, it should be noted that explaining highly-nonlinear ML models in a post-hoc fashion can lead to somewhat ambiguous insights or recommendations that violate common social norms such as ``do not penalize good attributes'' \cite{wang2020deontological}.

\paragraph{Perceptions of fairness and trustworthiness}
Due to the contested nature of the \emph{fairness} concept \cite{mulligan2019thing} as well as shortcomings of the typical AI/ML approaches to algorithmic fairness listed previously, researchers (primarily from HCI) have turned to more empirical and human-centric ways of assessing fairness and transparency of ADS.
For instance, \citet{binns2018s} and \citet{dodge2019explaining} compare human fairness perceptions of ADS for four \emph{distinct} post-hoc explanation styles.
Their works suggest differences in effectiveness of individual explanation styles---however, they also note that there does not seem to be a single best approach to explaining automated decisions.
\citet{lee2018understanding} compares perceptions of fairness and trustworthiness depending on whether the decision-maker is a person or an algorithm in the context of managerial decisions.
Their findings suggest that, among others, people perceive automated decisions as less fair and trustworthy for tasks that require typical human skills.
An interesting finding by \citet{lee2019procedural} suggests that fairness perceptions decline for some people when gaining an understanding of an algorithm if their personal fairness concepts differ from those of the algorithm.
Regarding trustworthiness, \citet{kizilcec2016much}, for instance, concludes that it is important to provide the right amount of transparency for optimal trust effects, as both too much and too little transparency can have undesirable effects.
One major weakness of related work is addressed by \citet{lee2021included}, who point out that prior studies have mostly recruited respondents from Amazon Mechanical Turk \cite{paolacci2010running}, which has predominantly white participants \cite{hitlin2016research}.
Moreover, it is often argued that one goal of explanations should be to facilitate positive perceptions (e.g., fairness and trustworthiness) towards ADS---which implicitly assumes that the respective ADS is fair and trustworthy, to begin with.
Examining which explanations allow for \emph{calibrated} perceptions is a significant research gap.
It will also be important to empirically measure novel ``flavors'' of fairness; for instance, based on established constructs from other disciplines like psychology.

\section{Research Objectives}\label{sec:objectives}
Based on research gaps identified in the previous section, I aim to contribute towards answering the following three main research questions through my doctoral thesis:
\begin{itemize}
    \item[\textbf{RQ1}] How do (potential) decision-subjects perceive consequential algorithmic decisions, primarily with respect to fairness; and how do explanations impact these perceptions?
    \item[\textbf{RQ2}] How and when can explanations enable (potential) decision-subjects to appropriately assess the quality (e.g., fairness) of ADS?
    \item[\textbf{RQ3}] Considering that in many real-world decision-making scenarios we do not have access to ground-truth labels, how can we design fair and transparent ADS that---unlike traditional ML approaches---do \emph{not} rest on this assumption?
\end{itemize}
I aim to investigate \textbf{RQ1} through an experimental study conducted with online study participants.
Contrary to existing work, I provide combinations of different explanation styles to (potential) decision-subjects, thereby simulating different levels of transparency.
I study perceptions of fairness and trustworthiness for the specific use case of bank lending, and I also analyze the moderating effects of people's AI literacy on their perceptions.
Based on findings from RQ1, I aim to analyze for \textbf{RQ2} whether people's perceptions are calibrated---meaning that they should be high \emph{if and only if} the underlying ADS is fair and trustworthy.
To that end, I aim to conduct a randomized online experiment as well.
Finally, regarding \textbf{RQ3}, I propose a paradigm shift from \emph{making accurate predictions} to \emph{making good decisions}.
Specifically, I aim to (a) make an algorithmic contribution to the fairness- and transparency-aware ADS toolbox; and (b) evaluate empirically how (potential) decision-subjects perceive this ``white-box'' artifact compared to traditional ML approaches with post-hoc explanations.

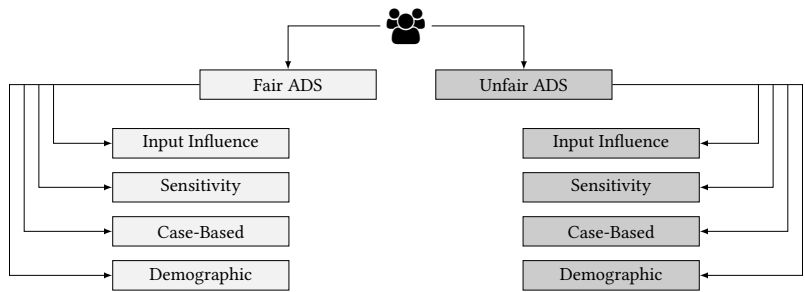
\begin{figure*}[t]
\centering
\resizebox{0.6\textwidth}{!}{
\begin{tikzpicture}  
  [scale=1] 
   
\node (n0) at (0,0) {\Huge \faUsers}; 

\node[style={rectangle, draw=black, fill=gray!10, minimum width=3cm, minimum height=0.5cm, text centered, anchor=east}] (n1) at (-0.5,-1) {Fair ADS}; 

\node[style={rectangle, draw=black, fill=gray!40, minimum width=3cm, minimum height=0.5cm, text centered, anchor=west}] (n2) at (0.5,-1) {Unfair ADS}; 


\node[style={rectangle, draw=black, fill=gray!10, minimum width=3cm, minimum height=0.5cm, text centered, anchor=west}] (n3) at (-5,-2) {Input Influence};

\node[style={rectangle, draw=black, fill=gray!10, minimum width=3cm, minimum height=0.5cm, text centered, anchor=west}] (n4) at (-5,-2.75) {Sensitivity}; 

\node[style={rectangle, draw=black, fill=gray!10, minimum width=3cm, minimum height=0.5cm, text centered, anchor=west}] (n5) at (-5,-3.5) {Case-Based};

\node[style={rectangle, draw=black, fill=gray!10, minimum width=3cm, minimum height=0.5cm, text centered, anchor=west}] (n6) at (-5,-4.25) {Demographic}; 


\node[style={rectangle, draw=black, fill=gray!40, minimum width=3cm, minimum height=0.5cm, text centered, anchor=east}] (n7) at (5,-2) {Input Influence};

\node[style={rectangle, draw=black, fill=gray!40, minimum width=3cm, minimum height=0.5cm, text centered, anchor=east}] (n8) at (5,-2.75) {Sensitivity}; 

\node[style={rectangle, draw=black, fill=gray!40, minimum width=3cm, minimum height=0.5cm, text centered, anchor=east}] (n9) at (5,-3.5) {Case-Based};

\node[style={rectangle, draw=black, fill=gray!40, minimum width=3cm, minimum height=0.5cm, text centered, anchor=east}] (n10) at (5,-4.25) {Demographic}; 


\draw[->, -latex] (n0) -| (n1);
\draw[->, -latex] (n0) -| (n2);

\draw[->, -latex] (-6,-1) |- (n3.west);
\draw[->, -latex] (-6.25,-1) |- (n4.west);
\draw[->, -latex] (-6.5,-1) |- (n5.west);
\draw[->, -latex] (n1.west) -- (-6.75,-1) |- (n6.west);

\draw[->, -latex] (6,-1) |- (n7.east);
\draw[->, -latex] (6.25,-1) |- (n8.east);
\draw[->, -latex] (6.5,-1) |- (n9.east);
\draw[->, -latex] (n2.east) -- (6.75,-1) |- (n10.east);

\end{tikzpicture}
}
\caption{Blueprint of preliminary study setup regarding \textbf{RQ2}.}
\label{fig:study_setup}
\Description{A blueprint of a randomized experiment, where half the study participants are shown output from a fair ADS, and the other half is shown output from an unfair ADS. Additionally, study participants are provided with one of four different explanation styles at random.}
\end{figure*}

\section{Results and Contributions to Date}\label{sec:results}
In this section, I briefly summarize main results and contributions of my dissertation-related academic work to date. An overview of peer-reviewed accepted and in-submission work is given in Table \ref{tab:my_work}.

\paragraph{Regarding \textbf{RQ1}}
For understanding people's perceptions towards ADS better, I have already conducted two studies.
The first one \cite{schoffer2021study,schoeffer2021there} examines the effects of ADS transparency on people's perceptions of informational fairness and trustworthiness.
To that end, I have designed and conducted a randomized online experiment around automated loan decisions with 400 participants recruited through Prolific \cite{palan2018prolific}.
Participants were assigned one of four transparency treatments (between-subject) and asked multiple questions based on established constructs from the information systems \cite{carter2005utilization} and organizational justice \cite{colquitt2001justice} literature.
Please refer to \cite{schoffer2021study} for samples of the employed questionnaires as well as the precise study setup including explanations.
Through analyzing group differences and estimating a full structural equation model (SEM), I found that an increase in transparency leads to an increase in both informational fairness and trustworthiness perceptions.
It is interesting to note, however, that informational fairness appears to act as a mediator between transparency and trustworthiness perceptions.
Additionally, I found evidence that people with higher AI literacy tend to perceive ADS as more informationally fair and trustworthy than people with little or no knowledge in this field.
In the second study \cite{schoeffer2021perceptions}, I have similarly conducted an online experiment (200 participants, recruited through Prolific) to compare perceptions between human-made and automated decisions in the context of lending.
Study participants were randomly assigned one of two treatments (human vs. automated), where both groups were shown \emph{identical} explanations about the decision-making logic---either employed by a human being or an ADS.
Interestingly, and contrary to some prior works' findings, I found that automated decisions are perceived as fairer than human-made decisions in the given context.
Based on an analysis of qualitative responses, it appears that people particularly appreciate the absence of subjectivity in ADS as well as their data-driven approach.
Again, please refer to the full publication \cite{schoeffer2021perceptions} for further details.

\paragraph{Regarding \textbf{RQ2}}
Based on the observations that (a) explanations can be misused to deceive certain groups of stakeholders (deliberately or unintentionally), and thus, (b) facilitating positive perceptions should \emph{not} be an unconditional goal of providing explanations, I have introduced the desideratum of \emph{appropriate fairness perceptions} \cite{schoeffer2021appropriate}.
This desideratum is satisfied if people's fairness perceptions towards an ADS are positive \emph{if and only if} the system is \emph{actually} fair.
A violation of this desideratum (e.g., when perceptions are positive despite the ADS issuing unfair decisions) likely indicates that people are not (sufficiently) enabled to appropriately assess the fairness of the respective ADS.
I further introduce in \cite{schoeffer2021appropriate} a novel study design to gauge the effectiveness of given explanation styles with respect to this desideratum.
I aim to spend a major share of the second half of my PhD program on conducting this study (see also Section \ref{sec:next_steps}).

\paragraph{Regarding \textbf{RQ3}}
As outlined in Section \ref{sec:related_work}, the general approach of in-processing fairness-aware ML techniques is to enforce certain statistical fairness constraints while maximizing for predictive performance.
This approach, however, is sub-optimal when ground-truth labels are unavailable \cite{kilbertus2017avoiding}.
In fact, in many real-world settings, we only have access to data with \emph{imperfect} labels, as the result of (potentially biased) human-made decisions.
In \cite{schoffer2021ranking}, I propose a novel ranking-based decision system that does \emph{not} learn to mimic biased decisions but (a) incorporates only \emph{useful} information from historical decisions, and (b) accounts for unwanted correlation between sensitive (e.g., gender) and legitimate features.
Specifically, I introduce a decision criterion based on weighted distances between individual data points and a so-called \emph{North Star}, which represents the (potentially hypothetical) observation with the highest possible qualification towards a given outcome.
By taking into account known (or easily identifiable) monotonic relationships between legitimate features and the outcome, the proposed decision criterion can be seen as a ``white-box'' approach that is readily interpretable by decision-subjects and allows for meaningful recourse (as opposed to many nonlinear ML models).
Through extensive experiments on synthetic and real-world data, I have shown that the proposed method is fair in the sense that it (a) assigns the desirable outcome to the most qualified individuals, and (b) removes the effect of stereotypes in decision-making, thereby outperforming traditional classification algorithms.
Finally, in \cite{schoffer2021ranking}, I have also shown theoretically that my method is consistent with the prominent individual fairness notion of FTA (see Section \ref{sec:related_work}).

\section{Next Steps and Outlook}\label{sec:next_steps}
Based on my results and contributions to date, most work remains to be done around \textbf{RQ2}.
In fact, after presenting and gathering feedback on the proposed idea and study setup of \cite{schoeffer2021appropriate} at CSCW 2021, I intend to design and carefully evaluate a suitable experiment.
A blueprint of the preliminary study setup is depicted in Figure \ref{fig:study_setup}.
Specifically, I aim to conduct a between-subject experiment where study participants are randomly shown either outcome from a fair or an unfair ADS.
Then, they are provided with a specific type of explanation regarding the decision-making logic.
The four explanation styles from Figure \ref{fig:study_setup} are taken from \citet{binns2018s}, but they could readily be replaced and/or extended by others.
Finally, I propose to measure fairness perceptions based on \citet{colquitt2015measuring}, and compare the results between the fair and unfair ADS treatments.
If perceptions are not substantially different, this indicates that a given explanation style does \emph{not} satisfy the desideratum of \emph{appropriate fairness perceptions}.
A key challenge in designing this experiment will be to align people's personal ``fairness concepts'' with the employed notion of fairness when constructing the fair and unfair ADS.
I aim to accomplish this through supplementary upstream experiments as well as through focusing on a specific community of people.
I generally think that this work will benefit tremendously from feedback of peers as well as more experienced HCI researchers.

Next, regarding \textbf{RQ3}, I plan to empirically evaluate my technical artifact \cite{schoffer2021ranking} with respect to people's perceptions and understanding.
I hypothesize that people appreciate the ``white-box'' approach---specifically, the fact that legitimate features are monotonically related to the final outcome; for instance, that higher GRE scores will never \emph{decrease} the chances of being admitted to grad school.
(Note that this is \emph{not} guaranteed for nonlinear ML models.)
I plan to test this hypothesis, among others, in a lab study setting.
Depending on the volume of this remaining work, I may have additional time to examine techniques for making my approach from \cite{schoffer2021ranking} applicable to regression settings as well.

With this dissertation at the intersection of HCI, computer science, and information systems, I aim to contribute towards solving issues of unfairness and non-transparency in algorithmic decision-making, from the perspective of the most vulnerable stakeholders: decision-subjects.
After finishing my PhD program, it is my utmost desire to continue conducting research on this important topic---either in a postdoctoral researcher role or as a researcher in industry.
My long-term career goal is to become a full professor.

\bibliographystyle{ACM-Reference-Format}
\bibliography{bibliography}


\begin{thebibliography}{58}


\ifx \showCODEN    \undefined \def \showCODEN     #1{\unskip}     \fi
\ifx \showDOI      \undefined \def \showDOI       #1{#1}\fi
\ifx \showISBNx    \undefined \def \showISBNx     #1{\unskip}     \fi
\ifx \showISBNxiii \undefined \def \showISBNxiii  #1{\unskip}     \fi
\ifx \showISSN     \undefined \def \showISSN      #1{\unskip}     \fi
\ifx \showLCCN     \undefined \def \showLCCN      #1{\unskip}     \fi
\ifx \shownote     \undefined \def \shownote      #1{#1}          \fi
\ifx \showarticletitle \undefined \def \showarticletitle #1{#1}   \fi
\ifx \showURL      \undefined \def \showURL       {\relax}        \fi
\providecommand\bibfield[2]{#2}
\providecommand\bibinfo[2]{#2}
\providecommand\natexlab[1]{#1}
\providecommand\showeprint[2][]{arXiv:#2}

\bibitem[\protect\citeauthoryear{Angwin, Larson, Mattu, and Kirchner}{Angwin
  et~al\mbox{.}}{2016}]%
        {angwin2016machine}
\bibfield{author}{\bibinfo{person}{Julia Angwin}, \bibinfo{person}{Jeff
  Larson}, \bibinfo{person}{Surya Mattu}, {and} \bibinfo{person}{Lauren
  Kirchner}.} \bibinfo{year}{2016}\natexlab{}.
\newblock \bibinfo{title}{Machine bias}.
\newblock
  \bibinfo{howpublished}{\url{https://www.propublica.org/article/machine-bias-risk-assessments-in-criminal-sentencing}}.
\newblock


\bibitem[\protect\citeauthoryear{Barocas, Hardt, and Narayanan}{Barocas
  et~al\mbox{.}}{2018}]%
        {barocas2018fairness}
\bibfield{author}{\bibinfo{person}{Solon Barocas}, \bibinfo{person}{Moritz
  Hardt}, {and} \bibinfo{person}{Arvind Narayanan}.}
  \bibinfo{year}{2018}\natexlab{}.
\newblock \showarticletitle{{Fairness and machine learning}}.
\newblock  (\bibinfo{year}{2018}).
\newblock
\urldef\tempurl%
\url{http://www.fairmlbook.org}
\showURL{%
\tempurl}


\bibitem[\protect\citeauthoryear{Binns, Van~Kleek, Veale, Lyngs, Zhao, and
  Shadbolt}{Binns et~al\mbox{.}}{2018}]%
        {binns2018s}
\bibfield{author}{\bibinfo{person}{Reuben Binns}, \bibinfo{person}{Max
  Van~Kleek}, \bibinfo{person}{Michael Veale}, \bibinfo{person}{Ulrik Lyngs},
  \bibinfo{person}{Jun Zhao}, {and} \bibinfo{person}{Nigel Shadbolt}.}
  \bibinfo{year}{2018}\natexlab{}.
\newblock \showarticletitle{'It's reducing a human being to a percentage'
  Perceptions of justice in algorithmic decisions}. In
  \bibinfo{booktitle}{\emph{Proceedings of the 2018 CHI Conference on Human
  Factors in Computing Systems}}. \bibinfo{pages}{1--14}.
\newblock


\bibitem[\protect\citeauthoryear{Buolamwini and Gebru}{Buolamwini and
  Gebru}{2018}]%
        {buolamwini2018gender}
\bibfield{author}{\bibinfo{person}{Joy Buolamwini} {and}
  \bibinfo{person}{Timnit Gebru}.} \bibinfo{year}{2018}\natexlab{}.
\newblock \showarticletitle{Gender shades: Intersectional accuracy disparities
  in commercial gender classification}. In \bibinfo{booktitle}{\emph{Conference
  on Fairness, Accountability, and Transparency}}. PMLR,
  \bibinfo{pages}{77--91}.
\newblock


\bibitem[\protect\citeauthoryear{Bussone, Stumpf, and O'Sullivan}{Bussone
  et~al\mbox{.}}{2015}]%
        {bussone2015role}
\bibfield{author}{\bibinfo{person}{Adrian Bussone}, \bibinfo{person}{Simone
  Stumpf}, {and} \bibinfo{person}{Dympna O'Sullivan}.}
  \bibinfo{year}{2015}\natexlab{}.
\newblock \showarticletitle{The role of explanations on trust and reliance in
  clinical decision support systems}. In \bibinfo{booktitle}{\emph{2015
  International Conference on Healthcare Informatics}}. IEEE,
  \bibinfo{pages}{160--169}.
\newblock


\bibitem[\protect\citeauthoryear{Calders, Kamiran, and Pechenizkiy}{Calders
  et~al\mbox{.}}{2009}]%
        {calders2009building}
\bibfield{author}{\bibinfo{person}{Toon Calders}, \bibinfo{person}{Faisal
  Kamiran}, {and} \bibinfo{person}{Mykola Pechenizkiy}.}
  \bibinfo{year}{2009}\natexlab{}.
\newblock \showarticletitle{Building classifiers with independency
  constraints}. In \bibinfo{booktitle}{\emph{2009 IEEE International Conference
  on Data Mining Workshops}}. IEEE, \bibinfo{pages}{13--18}.
\newblock


\bibitem[\protect\citeauthoryear{Carter and B{\'{e}}langer}{Carter and
  B{\'{e}}langer}{2005}]%
        {carter2005utilization}
\bibfield{author}{\bibinfo{person}{Lemuria Carter} {and}
  \bibinfo{person}{France B{\'{e}}langer}.} \bibinfo{year}{2005}\natexlab{}.
\newblock \showarticletitle{{The utilization of e-government services: Citizen
  trust, innovation and acceptance factors}}.
\newblock \bibinfo{journal}{\emph{Information Systems Journal}}
  \bibinfo{volume}{15}, \bibinfo{number}{1} (\bibinfo{year}{2005}),
  \bibinfo{pages}{5--25}.
\newblock


\bibitem[\protect\citeauthoryear{Chalfin, Danieli, Hillis, Jelveh, Luca,
  Ludwig, and Mullainathan}{Chalfin et~al\mbox{.}}{2016}]%
        {chalfin2016productivity}
\bibfield{author}{\bibinfo{person}{Aaron Chalfin}, \bibinfo{person}{Oren
  Danieli}, \bibinfo{person}{Andrew Hillis}, \bibinfo{person}{Zubin Jelveh},
  \bibinfo{person}{Michael Luca}, \bibinfo{person}{Jens Ludwig}, {and}
  \bibinfo{person}{Sendhil Mullainathan}.} \bibinfo{year}{2016}\natexlab{}.
\newblock \showarticletitle{{Productivity and selection of human capital with
  machine learning}}.
\newblock \bibinfo{journal}{\emph{American Economic Review}}
  \bibinfo{volume}{106}, \bibinfo{number}{5} (\bibinfo{year}{2016}),
  \bibinfo{pages}{124--127}.
\newblock


\bibitem[\protect\citeauthoryear{Chouldechova}{Chouldechova}{2017}]%
        {chouldechova2017fair}
\bibfield{author}{\bibinfo{person}{Alexandra Chouldechova}.}
  \bibinfo{year}{2017}\natexlab{}.
\newblock \showarticletitle{{Fair prediction with disparate impact: A study of
  bias in recidivism prediction instruments}}.
\newblock \bibinfo{journal}{\emph{Big Data}} \bibinfo{volume}{5},
  \bibinfo{number}{2} (\bibinfo{year}{2017}), \bibinfo{pages}{153--163}.
\newblock


\bibitem[\protect\citeauthoryear{Chromik, Eiband, V{\"o}lkel, and
  Buschek}{Chromik et~al\mbox{.}}{2019}]%
        {chromik2019dark}
\bibfield{author}{\bibinfo{person}{Michael Chromik}, \bibinfo{person}{Malin
  Eiband}, \bibinfo{person}{Sarah~Theres V{\"o}lkel}, {and}
  \bibinfo{person}{Daniel Buschek}.} \bibinfo{year}{2019}\natexlab{}.
\newblock \showarticletitle{Dark patterns of explainability, transparency, and
  user control for intelligent systems}. In \bibinfo{booktitle}{\emph{IUI
  Workshops}}, Vol.~\bibinfo{volume}{2327}.
\newblock


\bibitem[\protect\citeauthoryear{Colquitt, Conlon, Wesson, Porter, and
  Ng}{Colquitt et~al\mbox{.}}{2001}]%
        {colquitt2001justice}
\bibfield{author}{\bibinfo{person}{Jason~A Colquitt}, \bibinfo{person}{Donald~E
  Conlon}, \bibinfo{person}{Michael~J Wesson}, \bibinfo{person}{Christopher O
  L~H Porter}, {and} \bibinfo{person}{K~Yee Ng}.}
  \bibinfo{year}{2001}\natexlab{}.
\newblock \showarticletitle{{Justice at the millennium: A meta-analytic review
  of 25 years of organizational justice research}}.
\newblock \bibinfo{journal}{\emph{Journal of Applied Psychology}}
  \bibinfo{volume}{86}, \bibinfo{number}{3} (\bibinfo{year}{2001}),
  \bibinfo{pages}{425}.
\newblock


\bibitem[\protect\citeauthoryear{Colquitt and Rodell}{Colquitt and
  Rodell}{2015}]%
        {colquitt2015measuring}
\bibfield{author}{\bibinfo{person}{Jason~A Colquitt} {and}
  \bibinfo{person}{Jessica~B Rodell}.} \bibinfo{year}{2015}\natexlab{}.
\newblock \showarticletitle{{Measuring justice and fairness}}.
\newblock  (\bibinfo{year}{2015}).
\newblock


\bibitem[\protect\citeauthoryear{De-Arteaga, Fogliato, and
  Chouldechova}{De-Arteaga et~al\mbox{.}}{2020}]%
        {de2020case}
\bibfield{author}{\bibinfo{person}{Maria De-Arteaga}, \bibinfo{person}{Riccardo
  Fogliato}, {and} \bibinfo{person}{Alexandra Chouldechova}.}
  \bibinfo{year}{2020}\natexlab{}.
\newblock \showarticletitle{{A case for humans-in-the-loop: Decisions in the
  presence of erroneous algorithmic scores}}. In
  \bibinfo{booktitle}{\emph{Proceedings of the 2020 CHI Conference on Human
  Factors in Computing Systems}}. \bibinfo{pages}{1--12}.
\newblock


\bibitem[\protect\citeauthoryear{Dodge, Liao, Zhang, Bellamy, and Dugan}{Dodge
  et~al\mbox{.}}{2019}]%
        {dodge2019explaining}
\bibfield{author}{\bibinfo{person}{Jonathan Dodge}, \bibinfo{person}{Q~Vera
  Liao}, \bibinfo{person}{Yunfeng Zhang}, \bibinfo{person}{Rachel~KE Bellamy},
  {and} \bibinfo{person}{Casey Dugan}.} \bibinfo{year}{2019}\natexlab{}.
\newblock \showarticletitle{Explaining models: An empirical study of how
  explanations impact fairness judgment}. In
  \bibinfo{booktitle}{\emph{Proceedings of the 24th International Conference on
  Intelligent User Interfaces}}. \bibinfo{pages}{275--285}.
\newblock


\bibitem[\protect\citeauthoryear{Dwork, Hardt, Pitassi, Reingold, and
  Zemel}{Dwork et~al\mbox{.}}{2012}]%
        {dwork2012fairness}
\bibfield{author}{\bibinfo{person}{Cynthia Dwork}, \bibinfo{person}{Moritz
  Hardt}, \bibinfo{person}{Toniann Pitassi}, \bibinfo{person}{Omer Reingold},
  {and} \bibinfo{person}{Richard Zemel}.} \bibinfo{year}{2012}\natexlab{}.
\newblock \showarticletitle{Fairness through awareness}. In
  \bibinfo{booktitle}{\emph{Proceedings of the 3rd Innovations in Theoretical
  Computer Science Conference}}. \bibinfo{pages}{214--226}.
\newblock


\bibitem[\protect\citeauthoryear{Ehsan and Riedl}{Ehsan and Riedl}{2021}]%
        {ehsan2021explainability}
\bibfield{author}{\bibinfo{person}{Upol Ehsan} {and} \bibinfo{person}{Mark~O
  Riedl}.} \bibinfo{year}{2021}\natexlab{}.
\newblock \showarticletitle{Explainability pitfalls: Beyond dark patterns in
  explainable {AI}}.
\newblock \bibinfo{journal}{\emph{arXiv preprint arXiv:2109.12480}}
  (\bibinfo{year}{2021}).
\newblock


\bibitem[\protect\citeauthoryear{Eslami, Vaccaro, Lee, Elazari Bar~On, Gilbert,
  and Karahalios}{Eslami et~al\mbox{.}}{2019}]%
        {eslami2019user}
\bibfield{author}{\bibinfo{person}{Motahhare Eslami}, \bibinfo{person}{Kristen
  Vaccaro}, \bibinfo{person}{Min~Kyung Lee}, \bibinfo{person}{Amit Elazari
  Bar~On}, \bibinfo{person}{Eric Gilbert}, {and} \bibinfo{person}{Karrie
  Karahalios}.} \bibinfo{year}{2019}\natexlab{}.
\newblock \showarticletitle{User attitudes towards algorithmic opacity and
  transparency in online reviewing platforms}. In
  \bibinfo{booktitle}{\emph{Proceedings of the 2019 CHI Conference on Human
  Factors in Computing Systems}}. \bibinfo{pages}{1--14}.
\newblock


\bibitem[\protect\citeauthoryear{Feuerriegel, Dolata, and Schwabe}{Feuerriegel
  et~al\mbox{.}}{2020}]%
        {feuerriegel2020fair}
\bibfield{author}{\bibinfo{person}{Stefan Feuerriegel},
  \bibinfo{person}{Mateusz Dolata}, {and} \bibinfo{person}{Gerhard Schwabe}.}
  \bibinfo{year}{2020}\natexlab{}.
\newblock \showarticletitle{{Fair AI: Challenges and opportunities}}.
\newblock \bibinfo{journal}{\emph{Business \& Information Systems Engineering}}
   \bibinfo{volume}{62} (\bibinfo{year}{2020}), \bibinfo{pages}{379--384}.
\newblock


\bibitem[\protect\citeauthoryear{Goddard, Roudsari, and Wyatt}{Goddard
  et~al\mbox{.}}{2014}]%
        {Goddard2014}
\bibfield{author}{\bibinfo{person}{Kate Goddard}, \bibinfo{person}{Abdul
  Roudsari}, {and} \bibinfo{person}{Jeremy~C Wyatt}.}
  \bibinfo{year}{2014}\natexlab{}.
\newblock \showarticletitle{{Automation bias: Empirical results assessing
  influencing factors}}.
\newblock \bibinfo{journal}{\emph{International Journal of Medical
  Informatics}} \bibinfo{volume}{83}, \bibinfo{number}{5}
  (\bibinfo{year}{2014}), \bibinfo{pages}{368--375}.
\newblock
\showISSN{1386-5056}


\bibitem[\protect\citeauthoryear{Grote and Berens}{Grote and Berens}{2020}]%
        {grote2020ethics}
\bibfield{author}{\bibinfo{person}{Thomas Grote} {and} \bibinfo{person}{Philipp
  Berens}.} \bibinfo{year}{2020}\natexlab{}.
\newblock \showarticletitle{{On the ethics of algorithmic decision-making in
  healthcare}}.
\newblock \bibinfo{journal}{\emph{Journal of Medical Ethics}}
  \bibinfo{volume}{46}, \bibinfo{number}{3} (\bibinfo{year}{2020}),
  \bibinfo{pages}{205--211}.
\newblock


\bibitem[\protect\citeauthoryear{Hardt, Price, and Srebro}{Hardt
  et~al\mbox{.}}{2016}]%
        {hardt2016equality}
\bibfield{author}{\bibinfo{person}{Moritz Hardt}, \bibinfo{person}{Eric Price},
  {and} \bibinfo{person}{Nati Srebro}.} \bibinfo{year}{2016}\natexlab{}.
\newblock \showarticletitle{Equality of opportunity in supervised learning}. In
  \bibinfo{booktitle}{\emph{Advances in Neural Information Processing
  Systems}}. \bibinfo{pages}{3315--3323}.
\newblock


\bibitem[\protect\citeauthoryear{Harris and Davenport}{Harris and
  Davenport}{2005}]%
        {harris2005automated}
\bibfield{author}{\bibinfo{person}{Jeanne~G Harris} {and}
  \bibinfo{person}{Thomas~H Davenport}.} \bibinfo{year}{2005}\natexlab{}.
\newblock \showarticletitle{{Automated decision making comes of age}}.
\newblock \bibinfo{journal}{\emph{MIT Sloan Management Review}}
  \bibinfo{volume}{46}, \bibinfo{number}{4} (\bibinfo{year}{2005}),
  \bibinfo{pages}{2--10}.
\newblock


\bibitem[\protect\citeauthoryear{Heaven}{Heaven}{2020}]%
        {heaven2020predictive}
\bibfield{author}{\bibinfo{person}{Will~Douglas Heaven}.}
  \bibinfo{year}{2020}\natexlab{}.
\newblock \showarticletitle{{Predictive policing algorithms are racist. They
  need to be dismantled.}}
\newblock \bibinfo{journal}{\emph{MIT Technology Review}}
  (\bibinfo{year}{2020}).
\newblock


\bibitem[\protect\citeauthoryear{Hitlin}{Hitlin}{2016}]%
        {hitlin2016research}
\bibfield{author}{\bibinfo{person}{Paul Hitlin}.}
  \bibinfo{year}{2016}\natexlab{}.
\newblock \showarticletitle{Research in the crowdsourcing age: A case study}.
\newblock  (\bibinfo{year}{2016}).
\newblock


\bibitem[\protect\citeauthoryear{Imana, Korolova, and Heidemann}{Imana
  et~al\mbox{.}}{2021}]%
        {Imana21a}
\bibfield{author}{\bibinfo{person}{Basileal Imana}, \bibinfo{person}{Aleksandra
  Korolova}, {and} \bibinfo{person}{John Heidemann}.}
  \bibinfo{year}{2021}\natexlab{}.
\newblock \showarticletitle{Auditing for discrimination in algorithms
  delivering job ads}. In \bibinfo{booktitle}{\emph{Proceedings of the Web
  Conference 2021}}. \bibinfo{pages}{3767--3778}.
\newblock


\bibitem[\protect\citeauthoryear{Kilbertus, Rodriguez, Sch{\"o}lkopf, Muandet,
  and Valera}{Kilbertus et~al\mbox{.}}{2020}]%
        {kilbertus2020fair}
\bibfield{author}{\bibinfo{person}{Niki Kilbertus},
  \bibinfo{person}{Manuel~Gomez Rodriguez}, \bibinfo{person}{Bernhard
  Sch{\"o}lkopf}, \bibinfo{person}{Krikamol Muandet}, {and}
  \bibinfo{person}{Isabel Valera}.} \bibinfo{year}{2020}\natexlab{}.
\newblock \showarticletitle{Fair decisions despite imperfect predictions}. In
  \bibinfo{booktitle}{\emph{International Conference on Artificial Intelligence
  and Statistics}}. \bibinfo{pages}{277--287}.
\newblock


\bibitem[\protect\citeauthoryear{Kilbertus, Rojas-Carulla, Parascandolo, Hardt,
  Janzing, and Sch{\"o}lkopf}{Kilbertus et~al\mbox{.}}{2017}]%
        {kilbertus2017avoiding}
\bibfield{author}{\bibinfo{person}{Niki Kilbertus}, \bibinfo{person}{Mateo
  Rojas-Carulla}, \bibinfo{person}{Giambattista Parascandolo},
  \bibinfo{person}{Moritz Hardt}, \bibinfo{person}{Dominik Janzing}, {and}
  \bibinfo{person}{Bernhard Sch{\"o}lkopf}.} \bibinfo{year}{2017}\natexlab{}.
\newblock \showarticletitle{Avoiding discrimination through causal reasoning}.
  In \bibinfo{booktitle}{\emph{Advances in Neural Information Processing
  Systems}}. \bibinfo{pages}{656--666}.
\newblock


\bibitem[\protect\citeauthoryear{Kizilcec}{Kizilcec}{2016}]%
        {kizilcec2016much}
\bibfield{author}{\bibinfo{person}{Ren{\'e}~F Kizilcec}.}
  \bibinfo{year}{2016}\natexlab{}.
\newblock \showarticletitle{How much information? Effects of transparency on
  trust in an algorithmic interface}. In \bibinfo{booktitle}{\emph{Proceedings
  of the 2016 CHI Conference on Human Factors in Computing Systems}}.
  \bibinfo{pages}{2390--2395}.
\newblock


\bibitem[\protect\citeauthoryear{Kleinberg, Mullainathan, and
  Raghavan}{Kleinberg et~al\mbox{.}}{2016}]%
        {kleinberg2016inherent}
\bibfield{author}{\bibinfo{person}{Jon Kleinberg}, \bibinfo{person}{Sendhil
  Mullainathan}, {and} \bibinfo{person}{Manish Raghavan}.}
  \bibinfo{year}{2016}\natexlab{}.
\newblock \showarticletitle{Inherent trade-offs in the fair determination of
  risk scores}.
\newblock \bibinfo{journal}{\emph{arXiv preprint arXiv:1609.05807}}
  (\bibinfo{year}{2016}).
\newblock


\bibitem[\protect\citeauthoryear{Koivunen, Olsson, Olshannikova, and
  Lindberg}{Koivunen et~al\mbox{.}}{2019}]%
        {koivunen2019understanding}
\bibfield{author}{\bibinfo{person}{Sami Koivunen}, \bibinfo{person}{Thomas
  Olsson}, \bibinfo{person}{Ekaterina Olshannikova}, {and} \bibinfo{person}{Aki
  Lindberg}.} \bibinfo{year}{2019}\natexlab{}.
\newblock \showarticletitle{{Understanding decision-making in recruitment:
  Opportunities and challenges for information technology}}.
\newblock \bibinfo{journal}{\emph{Proceedings of the ACM on Human-Computer
  Interaction}} \bibinfo{volume}{3}, \bibinfo{number}{GROUP}
  (\bibinfo{year}{2019}), \bibinfo{pages}{1--22}.
\newblock


\bibitem[\protect\citeauthoryear{Kulesza, Stumpf, Burnett, Yang, Kwan, and
  Wong}{Kulesza et~al\mbox{.}}{2013}]%
        {kulesza2013too}
\bibfield{author}{\bibinfo{person}{Todd Kulesza}, \bibinfo{person}{Simone
  Stumpf}, \bibinfo{person}{Margaret Burnett}, \bibinfo{person}{Sherry Yang},
  \bibinfo{person}{Irwin Kwan}, {and} \bibinfo{person}{Weng-Keen Wong}.}
  \bibinfo{year}{2013}\natexlab{}.
\newblock \showarticletitle{Too much, too little, or just right? Ways
  explanations impact end users' mental models}. In
  \bibinfo{booktitle}{\emph{2013 IEEE Symposium on Visual Languages and Human
  Centric Computing}}. IEEE, \bibinfo{pages}{3--10}.
\newblock


\bibitem[\protect\citeauthoryear{Kuncel, Klieger, and Ones}{Kuncel
  et~al\mbox{.}}{2014}]%
        {kuncel2014hiring}
\bibfield{author}{\bibinfo{person}{Nathan~R Kuncel}, \bibinfo{person}{David~M
  Klieger}, {and} \bibinfo{person}{Deniz~S Ones}.}
  \bibinfo{year}{2014}\natexlab{}.
\newblock \showarticletitle{{In hiring, algorithms beat instinct}}.
\newblock \bibinfo{journal}{\emph{Harvard Business Review}}
  (\bibinfo{year}{2014}).
\newblock


\bibitem[\protect\citeauthoryear{Lakkaraju, Kleinberg, Leskovec, Ludwig, and
  Mullainathan}{Lakkaraju et~al\mbox{.}}{2017}]%
        {lakkaraju2017selective}
\bibfield{author}{\bibinfo{person}{Himabindu Lakkaraju}, \bibinfo{person}{Jon
  Kleinberg}, \bibinfo{person}{Jure Leskovec}, \bibinfo{person}{Jens Ludwig},
  {and} \bibinfo{person}{Sendhil Mullainathan}.}
  \bibinfo{year}{2017}\natexlab{}.
\newblock \showarticletitle{The selective labels problem: Evaluating
  algorithmic predictions in the presence of unobservables}. In
  \bibinfo{booktitle}{\emph{Proceedings of the 23rd ACM SIGKDD International
  Conference on Knowledge Discovery and Data Mining}}.
  \bibinfo{pages}{275--284}.
\newblock


\bibitem[\protect\citeauthoryear{Langer, Baum, K{\"o}nig, H{\"a}hne, Oster, and
  Speith}{Langer et~al\mbox{.}}{2021a}]%
        {langer2021spare}
\bibfield{author}{\bibinfo{person}{Markus Langer}, \bibinfo{person}{Kevin
  Baum}, \bibinfo{person}{Cornelius~J K{\"o}nig}, \bibinfo{person}{Viviane
  H{\"a}hne}, \bibinfo{person}{Daniel Oster}, {and} \bibinfo{person}{Timo
  Speith}.} \bibinfo{year}{2021}\natexlab{a}.
\newblock \showarticletitle{Spare me the details: How the type of information
  about automated interviews influences applicant reactions}.
\newblock \bibinfo{journal}{\emph{International Journal of Selection and
  Assessment}} (\bibinfo{year}{2021}).
\newblock


\bibitem[\protect\citeauthoryear{Langer, Oster, Speith, Hermanns, K{\"a}stner,
  Schmidt, Sesing, and Baum}{Langer et~al\mbox{.}}{2021b}]%
        {langer2021we}
\bibfield{author}{\bibinfo{person}{Markus Langer}, \bibinfo{person}{Daniel
  Oster}, \bibinfo{person}{Timo Speith}, \bibinfo{person}{Holger Hermanns},
  \bibinfo{person}{Lena K{\"a}stner}, \bibinfo{person}{Eva Schmidt},
  \bibinfo{person}{Andreas Sesing}, {and} \bibinfo{person}{Kevin Baum}.}
  \bibinfo{year}{2021}\natexlab{b}.
\newblock \showarticletitle{What do we want from explainable artificial
  intelligence (XAI)?---A stakeholder perspective on XAI and a conceptual model
  guiding interdisciplinary XAI research}.
\newblock \bibinfo{journal}{\emph{Artificial Intelligence}}
  \bibinfo{volume}{296} (\bibinfo{year}{2021}), \bibinfo{pages}{103473}.
\newblock


\bibitem[\protect\citeauthoryear{Lee}{Lee}{2018}]%
        {lee2018understanding}
\bibfield{author}{\bibinfo{person}{Min~Kyung Lee}.}
  \bibinfo{year}{2018}\natexlab{}.
\newblock \showarticletitle{{Understanding perception of algorithmic decisions:
  Fairness, trust, and emotion in response to algorithmic management}}.
\newblock \bibinfo{journal}{\emph{Big Data \& Society}} \bibinfo{volume}{5},
  \bibinfo{number}{1} (\bibinfo{year}{2018}), \bibinfo{pages}{1--16}.
\newblock


\bibitem[\protect\citeauthoryear{Lee, Jain, Cha, Ojha, and Kusbit}{Lee
  et~al\mbox{.}}{2019}]%
        {lee2019procedural}
\bibfield{author}{\bibinfo{person}{Min~Kyung Lee}, \bibinfo{person}{Anuraag
  Jain}, \bibinfo{person}{Hea~Jin Cha}, \bibinfo{person}{Shashank Ojha}, {and}
  \bibinfo{person}{Daniel Kusbit}.} \bibinfo{year}{2019}\natexlab{}.
\newblock \showarticletitle{{Procedural justice in algorithmic fairness:
  Leveraging transparency and outcome control for fair algorithmic mediation}}.
\newblock \bibinfo{journal}{\emph{Proceedings of the ACM on Human-Computer
  Interaction}} \bibinfo{volume}{3}, \bibinfo{number}{CSCW}
  (\bibinfo{year}{2019}), \bibinfo{pages}{182:1--182:26}.
\newblock


\bibitem[\protect\citeauthoryear{Lee and Rich}{Lee and Rich}{2021}]%
        {lee2021included}
\bibfield{author}{\bibinfo{person}{Min~Kyung Lee} {and}
  \bibinfo{person}{Katherine Rich}.} \bibinfo{year}{2021}\natexlab{}.
\newblock \showarticletitle{Who is included in human perceptions of AI? Trust
  and perceived fairness around healthcare AI and cultural mistrust}. In
  \bibinfo{booktitle}{\emph{Proceedings of the 2021 CHI Conference on Human
  Factors in Computing Systems}}. \bibinfo{pages}{1--14}.
\newblock


\bibitem[\protect\citeauthoryear{Lepri, Staiano, Sangokoya, Letouz{\'{e}}, and
  Oliver}{Lepri et~al\mbox{.}}{2017}]%
        {lepri2017tyranny}
\bibfield{author}{\bibinfo{person}{Bruno Lepri}, \bibinfo{person}{Jacopo
  Staiano}, \bibinfo{person}{David Sangokoya}, \bibinfo{person}{Emmanuel
  Letouz{\'{e}}}, {and} \bibinfo{person}{Nuria Oliver}.}
  \bibinfo{year}{2017}\natexlab{}.
\newblock \showarticletitle{{The tyranny of data? The bright and dark sides of
  data-driven decision-making for social good}}.
\newblock In \bibinfo{booktitle}{\emph{Transparent Data Mining for Big and
  Small Data}}. \bibinfo{publisher}{Springer}, \bibinfo{pages}{3--24}.
\newblock


\bibitem[\protect\citeauthoryear{Lewis and Mack}{Lewis and Mack}{1982}]%
        {lewis1982role}
\bibfield{author}{\bibinfo{person}{Clayton Lewis} {and} \bibinfo{person}{Robert
  Mack}.} \bibinfo{year}{1982}\natexlab{}.
\newblock \showarticletitle{{The role of abduction in learning to use a
  computer system}}.
\newblock  (\bibinfo{year}{1982}).
\newblock


\bibitem[\protect\citeauthoryear{Lundberg and Lee}{Lundberg and Lee}{2017}]%
        {lundberg2017unified}
\bibfield{author}{\bibinfo{person}{Scott~M Lundberg} {and}
  \bibinfo{person}{Su-In Lee}.} \bibinfo{year}{2017}\natexlab{}.
\newblock \showarticletitle{A unified approach to interpreting model
  predictions}. In \bibinfo{booktitle}{\emph{Proceedings of the 31st
  International Conference on Neural Information Processing Systems}}.
  \bibinfo{pages}{4768--4777}.
\newblock


\bibitem[\protect\citeauthoryear{Mathur, Acar, Friedman, Lucherini, Mayer,
  Chetty, and Narayanan}{Mathur et~al\mbox{.}}{2019}]%
        {mathur2019dark}
\bibfield{author}{\bibinfo{person}{Arunesh Mathur}, \bibinfo{person}{Gunes
  Acar}, \bibinfo{person}{Michael~J Friedman}, \bibinfo{person}{Elena
  Lucherini}, \bibinfo{person}{Jonathan Mayer}, \bibinfo{person}{Marshini
  Chetty}, {and} \bibinfo{person}{Arvind Narayanan}.}
  \bibinfo{year}{2019}\natexlab{}.
\newblock \showarticletitle{Dark patterns at scale: Findings from a crawl of
  11K shopping websites}.
\newblock \bibinfo{journal}{\emph{Proceedings of the ACM on Human-Computer
  Interaction}} \bibinfo{volume}{3}, \bibinfo{number}{CSCW}
  (\bibinfo{year}{2019}), \bibinfo{pages}{1--32}.
\newblock


\bibitem[\protect\citeauthoryear{Mehrabi, Morstatter, Saxena, Lerman, and
  Galstyan}{Mehrabi et~al\mbox{.}}{2021}]%
        {mehrabi2021survey}
\bibfield{author}{\bibinfo{person}{Ninareh Mehrabi}, \bibinfo{person}{Fred
  Morstatter}, \bibinfo{person}{Nripsuta Saxena}, \bibinfo{person}{Kristina
  Lerman}, {and} \bibinfo{person}{Aram Galstyan}.}
  \bibinfo{year}{2021}\natexlab{}.
\newblock \showarticletitle{A survey on bias and fairness in machine learning}.
\newblock \bibinfo{journal}{\emph{ACM Computing Surveys (CSUR)}}
  \bibinfo{volume}{54}, \bibinfo{number}{6} (\bibinfo{year}{2021}),
  \bibinfo{pages}{1--35}.
\newblock


\bibitem[\protect\citeauthoryear{Mulligan, Kroll, Kohli, and Wong}{Mulligan
  et~al\mbox{.}}{2019}]%
        {mulligan2019thing}
\bibfield{author}{\bibinfo{person}{Deirdre~K Mulligan},
  \bibinfo{person}{Joshua~A Kroll}, \bibinfo{person}{Nitin Kohli}, {and}
  \bibinfo{person}{Richmond~Y Wong}.} \bibinfo{year}{2019}\natexlab{}.
\newblock \showarticletitle{This thing called fairness: Disciplinary confusion
  realizing a value in technology}.
\newblock \bibinfo{journal}{\emph{Proceedings of the ACM on Human-Computer
  Interaction}} \bibinfo{volume}{3}, \bibinfo{number}{CSCW}
  (\bibinfo{year}{2019}), \bibinfo{pages}{1--36}.
\newblock


\bibitem[\protect\citeauthoryear{Newell and Marabelli}{Newell and
  Marabelli}{2015}]%
        {Newell2015}
\bibfield{author}{\bibinfo{person}{Sue Newell} {and} \bibinfo{person}{Marco
  Marabelli}.} \bibinfo{year}{2015}\natexlab{}.
\newblock \showarticletitle{{Strategic opportunities (and challenges) of
  algorithmic decision-making: A call for action on the long-term societal
  effects of ‘datification'}}.
\newblock \bibinfo{journal}{\emph{The Journal of Strategic Information
  Systems}} \bibinfo{volume}{24}, \bibinfo{number}{1} (\bibinfo{year}{2015}),
  \bibinfo{pages}{3--14}.
\newblock


\bibitem[\protect\citeauthoryear{Palan and Schitter}{Palan and
  Schitter}{2018}]%
        {palan2018prolific}
\bibfield{author}{\bibinfo{person}{Stefan Palan} {and}
  \bibinfo{person}{Christian Schitter}.} \bibinfo{year}{2018}\natexlab{}.
\newblock \showarticletitle{{Prolific.ac---A subject pool for online
  experiments}}.
\newblock \bibinfo{journal}{\emph{Journal of Behavioral and Experimental
  Finance}}  \bibinfo{volume}{17} (\bibinfo{year}{2018}),
  \bibinfo{pages}{22--27}.
\newblock


\bibitem[\protect\citeauthoryear{Paolacci, Chandler, and Ipeirotis}{Paolacci
  et~al\mbox{.}}{2010}]%
        {paolacci2010running}
\bibfield{author}{\bibinfo{person}{Gabriele Paolacci}, \bibinfo{person}{Jesse
  Chandler}, {and} \bibinfo{person}{Panagiotis~G Ipeirotis}.}
  \bibinfo{year}{2010}\natexlab{}.
\newblock \showarticletitle{Running experiments on Amazon Mechanical Turk}.
\newblock \bibinfo{journal}{\emph{Judgment and Decision making}}
  \bibinfo{volume}{5}, \bibinfo{number}{5} (\bibinfo{year}{2010}),
  \bibinfo{pages}{411--419}.
\newblock


\bibitem[\protect\citeauthoryear{Ribeiro, Singh, and Guestrin}{Ribeiro
  et~al\mbox{.}}{2016}]%
        {ribeiro2016should}
\bibfield{author}{\bibinfo{person}{Marco~Tulio Ribeiro},
  \bibinfo{person}{Sameer Singh}, {and} \bibinfo{person}{Carlos Guestrin}.}
  \bibinfo{year}{2016}\natexlab{}.
\newblock \showarticletitle{``Why should I trust you?'' Explaining the
  predictions of any classifier}. In \bibinfo{booktitle}{\emph{Proceedings of
  the 22nd ACM SIGKDD International Conference on Knowledge Discovery and Data
  Mining}}. \bibinfo{pages}{1135--1144}.
\newblock


\bibitem[\protect\citeauthoryear{Satariano}{Satariano}{2020}]%
        {satariano2020british}
\bibfield{author}{\bibinfo{person}{Adam Satariano}.}
  \bibinfo{year}{2020}\natexlab{}.
\newblock \showarticletitle{{British grading debacle shows pitfalls of
  automating government}}.
\newblock \bibinfo{journal}{\emph{The New York Times}} (\bibinfo{year}{2020}).
\newblock
\urldef\tempurl%
\url{https://www.nytimes.com/2020/08/20/world/europe/uk-england-grading-algorithm.html}
\showURL{%
\tempurl}


\bibitem[\protect\citeauthoryear{Schoeffer and Kuehl}{Schoeffer and
  Kuehl}{2021}]%
        {schoeffer2021appropriate}
\bibfield{author}{\bibinfo{person}{Jakob Schoeffer} {and}
  \bibinfo{person}{Niklas Kuehl}.} \bibinfo{year}{2021}\natexlab{}.
\newblock \showarticletitle{Appropriate fairness perceptions? On the
  effectiveness of explanations in enabling people to assess the fairness of
  automated decision systems}. In \bibinfo{booktitle}{\emph{Companion
  Publication of the 24th ACM Conference on Computer Supported Cooperative Work
  and Social Computing (CSCW ’21 Companion)}}.
\newblock


\bibitem[\protect\citeauthoryear{Schoeffer, Kuehl, and Valera}{Schoeffer
  et~al\mbox{.}}{2021a}]%
        {schoffer2021ranking}
\bibfield{author}{\bibinfo{person}{Jakob Schoeffer}, \bibinfo{person}{Niklas
  Kuehl}, {and} \bibinfo{person}{Isabel Valera}.}
  \bibinfo{year}{2021}\natexlab{a}.
\newblock \showarticletitle{A ranking approach to fair classification}. In
  \bibinfo{booktitle}{\emph{COMPASS '21: Proceedings of the 4th ACM SIGCAS
  Conference on Computing and Sustainable Societies}}.
\newblock


\bibitem[\protect\citeauthoryear{Schoeffer, Machowski, and Kuehl}{Schoeffer
  et~al\mbox{.}}{2021b}]%
        {schoeffer2021perceptions}
\bibfield{author}{\bibinfo{person}{Jakob Schoeffer}, \bibinfo{person}{Yvette
  Machowski}, {and} \bibinfo{person}{Niklas Kuehl}.}
  \bibinfo{year}{2021}\natexlab{b}.
\newblock \showarticletitle{Perceptions of fairness and trustworthiness based
  on explanations in human vs. automated decision-making}. In
  \bibinfo{booktitle}{\emph{55th Hawaii International Conference on System
  Sciences 2022 (HICSS-55)}}.
\newblock


\bibitem[\protect\citeauthoryear{Schoeffer, Machowski, and Kuehl}{Schoeffer
  et~al\mbox{.}}{2021c}]%
        {schoffer2021study}
\bibfield{author}{\bibinfo{person}{Jakob Schoeffer}, \bibinfo{person}{Yvette
  Machowski}, {and} \bibinfo{person}{Niklas Kuehl}.}
  \bibinfo{year}{2021}\natexlab{c}.
\newblock \showarticletitle{A study on fairness and trust perceptions in
  automated decision making}. In \bibinfo{booktitle}{\emph{Joint Proceedings of
  the ACM IUI 2021 Workshops}}.
\newblock


\bibitem[\protect\citeauthoryear{Schoeffer, Machowski, and Kuehl}{Schoeffer
  et~al\mbox{.}}{2021d}]%
        {schoeffer2021there}
\bibfield{author}{\bibinfo{person}{Jakob Schoeffer}, \bibinfo{person}{Yvette
  Machowski}, {and} \bibinfo{person}{Niklas Kuehl}.}
  \bibinfo{year}{2021}\natexlab{d}.
\newblock \showarticletitle{``There is not enough information'': On the effects
  of transparency on perceptions of informational fairness and trustworthiness
  in automated decision making}.
\newblock \bibinfo{journal}{\emph{Preprint}} (\bibinfo{year}{2021}).
\newblock


\bibitem[\protect\citeauthoryear{Srivastava, Heidari, and Krause}{Srivastava
  et~al\mbox{.}}{2019}]%
        {srivastava2019mathematical}
\bibfield{author}{\bibinfo{person}{Megha Srivastava}, \bibinfo{person}{Hoda
  Heidari}, {and} \bibinfo{person}{Andreas Krause}.}
  \bibinfo{year}{2019}\natexlab{}.
\newblock \showarticletitle{Mathematical notions vs. human perception of
  fairness: A descriptive approach to fairness for machine learning}. In
  \bibinfo{booktitle}{\emph{Proceedings of the 25th ACM SIGKDD International
  Conference on Knowledge Discovery \& Data Mining}}.
  \bibinfo{pages}{2459--2468}.
\newblock


\bibitem[\protect\citeauthoryear{Townson}{Townson}{2020}]%
        {townson2020ai}
\bibfield{author}{\bibinfo{person}{Sian Townson}.}
  \bibinfo{year}{2020}\natexlab{}.
\newblock \showarticletitle{{AI can make bank loans more fair}}.
\newblock \bibinfo{journal}{\emph{Harvard Business Review}}
  (\bibinfo{year}{2020}).
\newblock


\bibitem[\protect\citeauthoryear{Triberti, Durosini, and Pravettoni}{Triberti
  et~al\mbox{.}}{2020}]%
        {triberti2020third}
\bibfield{author}{\bibinfo{person}{Stefano Triberti}, \bibinfo{person}{Ilaria
  Durosini}, {and} \bibinfo{person}{Gabriella Pravettoni}.}
  \bibinfo{year}{2020}\natexlab{}.
\newblock \showarticletitle{{A ``third wheel'' effect in health decision making
  involving artificial entities: A psychological perspective}}.
\newblock \bibinfo{journal}{\emph{Frontiers in Public Health}}
  \bibinfo{volume}{8} (\bibinfo{year}{2020}).
\newblock


\bibitem[\protect\citeauthoryear{Wang and Gupta}{Wang and Gupta}{2020}]%
        {wang2020deontological}
\bibfield{author}{\bibinfo{person}{Serena Wang} {and} \bibinfo{person}{Maya
  Gupta}.} \bibinfo{year}{2020}\natexlab{}.
\newblock \showarticletitle{Deontological ethics by monotonicity shape
  constraints}. In \bibinfo{booktitle}{\emph{International Conference on
  Artificial Intelligence and Statistics}}. PMLR, \bibinfo{pages}{2043--2054}.
\newblock


\end{thebibliography}

\end{document}